\pdfoutput=1

\documentclass[11pt]{article}

\usepackage[final]{EMNLP2022}

\usepackage{times}
\usepackage{latexsym}

\usepackage[T1]{fontenc}

\usepackage[utf8]{inputenc}

\usepackage{latexsym}
\usepackage{soul}
\usepackage{multicol}
\usepackage[many]{tcolorbox}
\usepackage{stfloats,framed}
\usepackage{lipsum}
\usepackage{amsmath}
\usepackage{multirow}
\usepackage{float}
\usepackage{booktabs}

\usepackage{amsfonts}

\newcommand{\hlc}[2][yellow]{{%
    \colorlet{foo}{#1}%
    \sethlcolor{foo}\hl{#2}}%
}
\usepackage{microtype}
\usepackage{subfigure}
\usepackage{inconsolata}
\usepackage{microtype}
\definecolor{ao}{rgb}{0.0, 0.5, 0.0}

\newcommand{\rom}[1]{\expandafter{\romannumeral #1\relax}}

\title{Neural Keyphrase Generation: Analysis and Evaluation}

\author{\textbf{Tuhin Kundu}$^\clubsuit$ \thanks{\;\; Work done at the University of Illinois at Chicago before Amazon} $\quad$ \textbf{Jishnu Ray Chowdhury}$^\spadesuit$ $\quad$
        \textbf{Cornelia Caragea}$^\spadesuit$\\
   $^\clubsuit$Amazon $\quad$ 
   $^\spadesuit$ Computer Science, University of Illinois at Chicago \\
  {\tt \{jraych2,cornelia\}@uic.edu} \\
  {\tt tuhinkundu@outlook.com} \\
}

\begin{document}
\maketitle
\begin{abstract}
Keyphrase generation aims at generating topical phrases from a given text either by copying from the original text (present keyphrases) or by producing new keyphrases (absent keyphrases) that capture the semantic meaning of the text. Encoder-decoder models are most widely used for this task because of their capabilities for absent keyphrase generation. However, there has been little to no analysis on the performance and behavior of such models for keyphrase generation. In this paper, we study various tendencies exhibited by three strong models: T5 (based on a pre-trained transformer), CatSeq-Transformer (a non-pretrained Transformer), and ExHiRD (based on a recurrent neural network). We analyze prediction confidence scores, model calibration, and the effect of token position on keyphrases generation. Moreover, we motivate and propose a novel metric framework, SoftKeyScore, to evaluate the similarity between two sets of keyphrases by using soft-scores to account for partial matching and semantic similarity. We find that SoftKeyScore is more suitable than the standard F$_{1}$ metric for evaluating two sets of given keyphrases.
\end{abstract}

\section{Introduction}

Keyphrase generation is the task of predicting a set of keyphrases from a given document that capture the core ideas and topics of the document. Among these keyphrases, some exist within the source document (present keyphrases), and some are absent from the document (absent keyphrases).  Keyphrases are widely used in various applications, such as document indexing and retrieval \cite{jones1999phrasier, boudin2020keyphrase}, document clustering \cite{hulth2006study}, and text summarization \cite{wang2013domain}. Hence, keyphrase generation is of great interest to the scientific community.

In recent years, neural encoder-decoder (seq2seq) models are adapted to generate both absent and present keyphrases \cite{meng2017deep}. Contemporary approaches \cite{yuan2020one, chan2019neural, chen2020exclusive} to keyphrase generation aim at autoregressively decoding a sequence of concatenated keyphrases from a given source document. Typically, these models are equipped with cross-attention \cite{luong2015effective, bahdanau2015neural} and a copy (or pointer) mechanism \cite{gu2016incorporating, see-etal-2017-get}. 
Although several variants and extensions of seq2seq models have been proposed to enhance keyphrase generation \cite{meng2017deep, yuan2020one, chan2019neural, swaminathan-etal-2020-preliminary, chen2020exclusive}, there have been limited attempts at deeper analysis on the tendencies of neural seq2seq models in this task. Moreover, the ubiquitous success of pre-trained models for a multitude of NLP tasks motivated us to also investigate and analyze the application of a pre-trained seq2seq models, T5 \cite{raffel2020exploring}, for keyphrase generation. Overall, we contrast the performance of T5 with a strong recurrent neural network (RNN) named (ExHiRD) \cite{chen2020exclusive} on different aspects of keyphrase generation and, to better disentangle the effect of pre-training, we also compare it against a Transformer-based seq2seq model (CatSeq-Transformer) trained from scratch (i.e., not pre-trained). Below, we discuss and motivate the factors of keyphrase generation models that we mainly analyze:

\vspace{1mm}
\noindent \textbf{Model Calibration and Uncertainty:}
In practical applications, it is often desirable to estimate the confidence of a model prediction to decide on whether that prediction can be used or not. Similarly in keyphrase generation, in principle, model confidence could be used to make different decisions (for example, ranking keyphrases after overgeneration, or mixing predictions of different models based on their confidence). However, before we can rely on the confidence estimated by a model (based on its prediction probabilities), we need to determine how well calibrated the model is. A well-calibrated model should generally ``know what it does not know'', which can be reflected by a strong alignment between its empirical likelihood (accuracy) and its probability estimates (confidence).
Thus, in this work, we measure and contrast the expected calibration errors (ECE) of all the considered models. To be able to measure ECE at the level of keyphrases, we propose a novel perplexity-based measure called {\em Keyphrase Perplexity} (KPP) with which we also analyze a model's own estimated confidence for predicting different keyphrase types. 

\vspace{1mm}
\noindent \textbf{Robustness to Positional Variance:}
Keyphrases in scientific-domain can often first appear early in the given document \cite{florescu-caragea-2017-positionrank}. Focusing only in the early areas of the given text can thus often be a useful heuristic to predict keyphrases (similar to lead bias in summarization). However, robust models should be able to predict keyphrases well no matter in which positions they occur. Thus, we also check for the variance of model performance on present keyphrases with varying positions of their first occurrence. 

\vspace{1mm}
\noindent \textbf{Evaluation with Partial-match-based F$_1$:}
Standardly, in the keyphrase generation and extraction literature, $F_1$ and $recall$ based metrics are used to evaluate the keyphrase prediction performance across different models (usually often accommodated with some top-k keyphrase selection policy for a ranked list of unique keyphrases). However, all such metrics are based on exact match scores between the stemmed versions of predicted keyphrases and gold ones. Such a strategy cannot account for partial matches or semantic similarity. For example, if the prediction is ``summarization model'' and the gold is ``summarization system'', despite both semantic similarity and partial matching, the score will be 0. These kind of minor deviations are ubiquitous in keyphrase generation yet they are harshly penalized by the “exact match” evaluation metrics. This phenomenon motivated us to propose a novel soft-scoring based evaluation paradigm, SoftKeyScore, that is specifically suited for evaluating sets (not sequence) of phrases. 

\vspace{-2mm}
\section{Methodology}





For our analysis, we consider three models: ExHiRD, CatSeq-Transformer, and T5. We chose ExHiRD because it is one of the strongest performing keyphrase generation architectures without relying on reinforcement learning or GANs. We chose T5 because applications of pre-trained Transformer-based models like T5 are becoming almost ubiquitous in NLP and T5 serves as a natural choice for keyphrase generation given its seq2seq architecture. We chose CatSeq-Transformer to show the effect of simply using a Transformer-based architecture over a specialized RNN-based one when both have no pre-training. 
All the models are trained on the concatenated sequence of target keyphrases as in \citet{yuan2020one}. 
Implementation details for the models are presented in Appendix \ref{eval}.
\vspace{1mm}

\noindent\textbf{ExHiRD:} ExHiRD \cite{chen2020exclusive} is an RNN-based seq2seq model with attention and copy-mechanism. It uses a hierarchical decoding strategy to address the hierarchical nature of a sequence of keyphrases, where each keyphrase is, in turn, a sub-sequence of words. ExHiRD also proposes exclusion mechanisms to improve the diversity of keyphrases generated 
and reduce duplication. 
\vspace{1mm}

\noindent \textbf{T5:} T5 \cite{raffel2020exploring} is a pre-trained seq2seq Transformer \cite{vaswani2017attention}, which is pre-trained on C4 corpus (a dataset with clean English text obtained by scraping the Web). The T5 architecture includes an encoder-decoder architecture with various layers of self-attention and cross attention.
We use {\fontfamily{qcr}\selectfont t5-base} model with 12 layers from the Transformers library \cite{wolf2020transformers}.
\vspace{1mm}

\noindent \textbf{CatSeq-Transformer:} As we discussed, CatSeq-Transformer is simply the vanilla Transformer model that is trained on keyphrase generation in the CatSeq paradigm \cite{yuan2020one} without prior pre-training.  
\vspace{1mm}

\noindent For the sake of brevity, we mainly present the results of comparing ExHiRD and T5 in the main paper. We still summarize our findings involving CatSeq-Transformer in the relevant sections of the main paper, but we put most of the details in Appendix \ref{section:vanilla_transformer_comparison}. Below, we provide the technical backgrounds for our different approaches to analyzing and evaluating the aforementioned models. 



\subsection{Model Calibration and Uncertainty}

As we discussed before, it is important to check how well calibrated a given model is. In this section, we first present a novel measure, {\em Keyphrase perplexity} (KPP), to estimate a model's confidence at the level of keyphrases and then we describe how we use KPP to estimate calibration.

\subsubsection{Keyphrase Perplexity}
We propose \textit{Keyphrase Perplexity} ($KPP$) to gauge model confidence on a particular predicted keyphrase. $KPP$ is rooted in the general concept of perplexity. Perplexity is a widely used metric for evaluating language models. For a sequence of tokens $w_{1:n} = {w_{1}, w_{2},...,w_{n}}$ of length $n$, perplexity is the inverse normalized probability $p$ of generating them and can be defined as: $PP(w_{1:n}) = p(w_{1},w_{2},...,w_{n})^{-1/n}$. For an auto-regressive decoder, the probability $p$ of the sequence can be factorized and reformulated as:

\vspace{-5mm}
\begin{equation}
PP(w_{1:n}) = \left(\prod_{i=1}^{n} p(w_{i} | w_1, w_2,\dots w_{i-1})\right)^{-1/n}
\label{standard_pp}
\end{equation}

\noindent However, note that in the widely used CatSeq framework \cite{yuan2020one}, a generated/decoded sequence is a concatenation of keyphrases.
The vanilla perplexity is only defined over the whole generated sequence and cannot be directly applied for subsequences (keyphrases) within the sequence. Thus, to get an estimate of the model confidence at the level of predicting individual keyphrases, we adapt the original perplexity and define keyphrase perplexity ($KPP$) as follows. Given a particular keyphrase represented as the sub-sequence $w_{j:k} = {w_{j}, w_{j+1},...,w_{k}}$ within the sequence $w_{1:n}$ $(1 \leq j \leq k\leq n)$ (representing a sequence of concatenated keyphrases), the KPP of that keyphrase ($w_{j:k}$) is defined as:

\vspace{-6mm}
\begin{equation}
KPP(w_{j:k}) = \left(\prod_{i=j}^{k} p(w_{i} | w_1, w_2,\dots w_{i-1})\right)^{-1/m}
\label{eq:KPP}
\end{equation}

\noindent where $m = k-j+1$ is the number of tokens in the keyphrase $w_{j:k}$. Essentially, for $KPP$, we simply use the conditional probabilities of tokens within the keyphrase $w_{j:k}$ under consideration.\footnote{One limitation of this $KPP$ formulation is that it does not negate the conditioning effect of previous keyphrases (included in sub-sequence $w_1$ to $w_{j-1}$ while measuring the $KPP$ of the keyphrase starting from $w_j$). However, removing this limitation is not straight-forward; so we take a naive assumption of treating the overall probabilities of keyphrases as independent of the other keyphrases. As such, our formulation is a form of ``quasi-perplexity" measure.} 
During our analysis, any probability of the form $p(w_{i} | w_1, w_2,\dots w_{i-1})$ indicates the predicted model probability for token $w_i$ given that tokens $w_1, w_2,\dots w_{i-1}$ have been already generated. 
We do not consider special tokens (e.g., keyphrase delimiters or end of sequence markers) as part of any keyphrase subsequence for $KPP$.
As in perplexity, a lower $KPP$  indicates a higher confidence in the prediction, whereas a higher $KPP$  indicates a lower confidence. 

\vspace{-1mm}



\subsubsection{Calibration}

Model calibration reflects the accuracy of model predictions as a function of its generated posterior probabilities. A calibrated model has alignment between its empirical likelihood (accuracy) and its probability estimates (confidence). For example, a calibrated model that has a confidence of $90\%$ while making predictions, would correctly predict 90 out of 100 possible samples. Formally, calibration models the joint distribution $P(Q,Y)$ over generated model probabilities $Q \in \mathbb{R}$ and labels $Y$. $P(Y=y|Q=q)=q$ signifies perfect calibration of a model \cite{guo2017calibration}.

Expected calibration error (ECE) is a popular measure of model miscalibration \cite{ece}. ECE is computed by partitioning the predictions according to their confidence estimates into $k$ bins (we set $k$=10) and summing up the weighted average of the absolute value of the difference between the accuracy and the average confidence of keyphrases in each bin. This can be formalized as:
\vspace{-1mm}
\begin{equation} \label{eq:ECE}
ECE = \sum_{i=1}^{k} \frac{|B_{i}|}{n} |acc(B_{i}) - confid(B_{i})|
\end{equation}
\vspace{-1mm}

\noindent Here $n$ is the number of total samples,  $|B_{i}|$ is the number of samples in bin $B_i$, $1 \leq i \leq k$, of $k$ bins. In our task, we compute $acc(B_i)$ as the fraction of accurately predicted keyphrases in bin $B_i$ and $confid(B_i)$ as the average confidence in bin $B_i$. We define confidence of a particular generated keyphrase as the inverse of its KPP ($KPP^{-1}$) that is, roughly, the length normalized product of posterior probabilities for the tokens of that keyphrase.

In addition to ECE, reliability diagrams depict the accuracy of the model as a function of the probability across the $k$ bins. 


\subsection{Robustness to Positional Variance}

To analyze the robustness of different models on detecting present keyphrases in different positions in the input document, we divide the input document into five sections (bins) with $20\%$ of characters in each, and binned the keyphrases appearing in them accordingly. In Table \ref{positiontable}, we see that the majority of gold labels for the present keyphrases in the scientific datasets that we consider are in the first section (bin) of the input sequence. Thus, there can be lead bias in scientific abstracts. If a model learns to overexploit this lead bias by mostly focusing on the early sections, it can miss important keyphrases that occur in later sections. We use the percentage of missed keyphrases in each bin for each of the considered models to check how their performance varies with the varying position of present keyphrases.

\begin{table}[t]
\small
\centering
\begin{tabular}{c|rrrrr}
\toprule
\multirow{2}{*}{Dataset} & \multicolumn{5}{c}{Document Section} \\ 
 & \multicolumn{1}{c}{1} & \multicolumn{1}{c}{2} & \multicolumn{1}{c}{3} & \multicolumn{1}{c}{4} & \multicolumn{1}{c}{5} \\ \midrule
Inspec & 1,326 & 845 & 686 & 602 & 173 \\
Krapivin & 706 & 206 & 182 & 159 & 59 \\
SemEval & 346 & 126 & 103 & 54 & 20 \\
KP20k & 39,571 & 9,865 & 8,313 & 6,317 & 1,704 \\ \bottomrule
\end{tabular}
\caption{Document Section (bin) 1 represents the first $20\%$ of all characters in the given text, Document Section 2 indicates the second $20\%$ portion and so on. We show the total number of gold present keyphrases in each of the five sections.}
\label{positiontable}
\end{table} 

\begin{table*}[h]
\resizebox{\textwidth}{!}{%
\small
\centering
\begin{tabular}{l|c|c|ccccc}
\toprule
\multicolumn{1}{c|}{Examples} & F$_1$@M & F$_{KMR}$@M &  \multicolumn{3}{c}{F$_{BERTScore}$@M} \\
\midrule
& & & DeBERTa & RoBERTA & SciBERT \\ 
\midrule
\begin{tabular}[c]{@{}l@{}}\textbf{Pred}: performance evaluation, information retrieval, web search engine\\ \textbf{Gold}: performance, information retrieval, world wide web, search engine\end{tabular} & 0.286 & 0.375 & 0.520 & 0.568 & 0.618 \\ \midrule
\begin{tabular}[c]{@{}l@{}}\textbf{Pred}: bgp, network engineering, routing protocols\\ \textbf{Gold}: routing, traffic engineering, modeling, bgp\end{tabular} & 0.286 & 0.500 & 0.538 & 0.549 & 0.671 \\ 
\midrule
\begin{tabular}[c]{@{}l@{}}\textbf{Pred}: pwarx identification, chiu's clustering algorithm, \\affine sub model estimation, hyperplane partitions \\ \textbf{Gold}: experimental validation, clustering, identification, hybrid systems, \\pwarx models, chiu's clustering technique
\end{tabular} & 0.000 & 0.083 & 0.234 & 0.260 & 0.493  \\ 
\bottomrule
\end{tabular}
}
\caption{{\bf Pred} represents set of predicted keyphrases from a sample input, {\bf Gold} represents the corresponding set of gold keyphrases. The table shows the corresponding evaluated values from both exact match-based F$_1$@M metric and our SoftKeyScores with different score functions for each pairs of keyphrase sets.}
\label{table:bertscore_examples}
\end{table*}

\subsection{Evaluation with Partial-match-based F$_1$}


Previously, we motivated the need for an evaluation metric that accounts for partial matches or semantic similarities between keyphrases. Here, we define the technical framework ({\em SoftKeyScore}) that we introduce as a solution for that need. 

Assume we have two sets $G = \{g_1, g_2, ..., g_{|G|}\}$ and $P = \{p_1, p_2, ..., p_{|P|}\}$. $G$ can be the set of gold keyphrases and $P$ can be the set of predicted keyphrases. Assume we also have some soft-scoring function $score(x,y)$ which takes two phrases ($x$ and $y$) as input and outputs a scalar $\in [0,1]$ to indicate the degree of match between $x$ and $y$. 
Given these elements, we propose the following evaluation framework:
\begin{align}
&P_{score} = \frac{1}{|P|} \cdot \sum_{p_i \in p} \underset{g_j \in G}{\max} \; score(p_i,g_j) \label{meta-p-score}\\
&R_{score} = \frac{1}{|G|} \cdot \sum_{g_j \in g} \underset{p_i \in P}{\max} \; score(p_i,g_j) \label{meta-r-score}\\
&F_{score} = 2 \cdot \frac{\cdot P_{score} \cdot R_{score}}{P_{score}+R_{score}} \label{meta-f-score}
\end{align}

\noindent Here, $F_{score}$ indicates the final result of SoftKeyScore. It is analogous to $F_1$; the difference is in how the precision and recall are computed. 
$P_{score}$ and $R_{score}$ are analogous to precision and recall, respectively (in fact, they can be considered as more relaxed forms of precision and recall). With a soft scoring function ($score$), however, one phrase $p_i$ in set $P$ can match with multiple phrases in set $G$. Thus, in Eqs. \ref{meta-p-score} and \ref{meta-r-score}, we use a greedy matching strategy where we choose the maximum matching score for any comparison between a phrase in one set to all phrases in the other set. This overall framework is very similar to the framework used for BERTScore \cite{zhang2019bertscore}. However, the crucial difference is that we are using a generic matching function to measure similarity between two \textit{sequences} (keyphrases) instead of two token embeddings. In fact, one of our proposed scoring functions (discussed below) uses BERTScore itself. 

SoftKeyScore is invariant to the order of phrases. This is suitable in our context of evaluating \textit{sets} of keyphrases. At the same time, by using the right $score$ function, we can account for the order among the words within phrases. Note that if we simply use an exact match \textit{score} function, SoftKeyScore reduces to the standard $F_1$. As such, SoftKeyScore can be considered as a generalization of the standard $F_1$ metric. Below we discuss two concrete instances of the $score$ function that we explore in our calculation of SoftKeyScore: Keyphrase Match Rate (KMR) score and BERTScore. Given two phrases, KMR relies purely on surface-string-level features whereas BERTScore can model their semantic similarity. More implementation details of this framework can be found in Appendix \ref{section:bertscore_appendix}. 
\vspace{1mm}

\noindent \textbf{Keyphrase Match Rate (KMR)} \quad We propose Keyphrase Match Rate ($KMR$) as the complement of Translation Error Rate (TER) \cite{snover2006study}. TER is a NLG metric based on edit-distance between two strings. 
In our work, we slightly modify the original TER score by adding pads to the shorter sequence (keyphrase) to keep the lengths of the two sequences under comparison equal. Pad tokens change some deletions to substitutions but that does not change the total edit cost since both have the same cost yet this strategy ensures that TER stays in $[0,1]$.
Given that we want to measure the similarity (not distance) between two keyphrases, we now formulate $KMR$ as: $1 - TER$. Like TER, KMR also ranges in $[0,1]$. 

\vspace{1mm}
\noindent \textbf{BERTScore} \quad BERTScore \cite{zhang2019bertscore} is a recently proposed metric for the evaluation of natural language generation models. BERTScore uses a similar method as described in Eqs. \ref{meta-p-score} - \ref{meta-f-score}, but with the following differences:
\begin{enumerate}
\vspace{-2mm}
    \item Instead of sets ($P$ and $G$), the original BERTScore evaluation is done on two sequences of tokens (prediction sequence and reference sequence). 
    \vspace{-2mm}
    \item Instead of phrases from some given sets, the equivalent of $score$ function in BERTScore compares contextualized token embeddings from the given sequences using dot-product.
\end{enumerate} 
\vspace{-2mm}
In our context, we use BERTScore as another instance of the $score$ function as described previously to measure the similarity between two phrases. BERTScore can take into account both partial matching and deeper semantic similarities between the two phrases. Note that if we \textit{just} use BERTScore replacing SoftKeyScore, the evaluation will no longer be invariant to the order of the keyphrases because of the use of contextualized embeddings over a ``sequence'' (it will no longer remain a set) of keyphrases.

\subsubsection{SoftKeyScores Examples}
In our notations, we specify that we are using a specific SoftKeyScore by writing $F$ followed by a subscript where the specific score function is written. For example, $F_{KMR}$ denotes SoftKeyScore based on KMR, and $F_{BERTScore}$ denotes SoftKeyScore based on BERTScore. We can then append $@M$ or $@5$ to the notation, just as in the traditional F$_1$ metrics, to denote how many prediction keyphrases are selected ($@M$ indicates all predictions are selected, whereas $@5$ indicates that the top $5$ ones are selected). Table 2 provides some concrete examples  with comparison between exact match based F$_1$@M and SoftKeyScore with different score functions when applied to two given sets of keyphrases. We use DeBERTa \cite{he2021deberta}, RoBERTa \cite{liu2020roberta}, and SciBERT \cite{beltagy2019scibert} as alternatve BERT models to compute BERTScore. Further details are in Appendix \ref{section:bertscore_appendix}. As we can see, the exact-match F1 metrics are quite low despite high similarities of the predictions and targets. In such cases, SoftKeyScore, can better fit our intuitions about similarity between sets of phrases.

\section{Experiments and Results}
\vspace{-2mm}
We select four widely used benchmarks for our experimentation: \textbf{KP20k} \cite{meng2017deep}, \textbf{Krapivin} \cite{krapivin2009large}, \textbf{Inspec} \cite{hulth2003improved} and \textbf{SemEval} \cite{kim2010semeval}. We use KP20k training set ($\sim$500,000 samples) for training our models. As test sets, we use the test sets available for each dataset 
for performance evaluation and analysis. 
Implementation details and evaluation metrics are in Appendices \ref{eval} and \ref{evaluation}. 

\vspace{1mm}
\noindent \textbf{Keyphrase Perplexity Analysis:}
We compare keyphrase perplexities ($KPP$) of both T5 and ExHiRD (using histograms) in Figure \ref{confidence}. In Appendix \ref{section:vanilla_transformer_comparison} (Figure \ref{fig:all_three_perplexities}), we also show these results for CatSeq-Transformers. Unsurprisingly, we find that all models have lower $KPP$ (thus, higher confidence) for present keyphrases than absent keyphrases (which are harder to learn to generate). However, T5 is substantially more confident about its present keyphrase predictions compared to ExHiRD. 
This appears to be the effect of pre-training in T5, because we find (\ref{fig:all_three_perplexities}) that the non-pretrained CatSeq-Transformers is generally the least confident. 

\begin{figure}[t]
 \center
  \includegraphics[width=\linewidth]{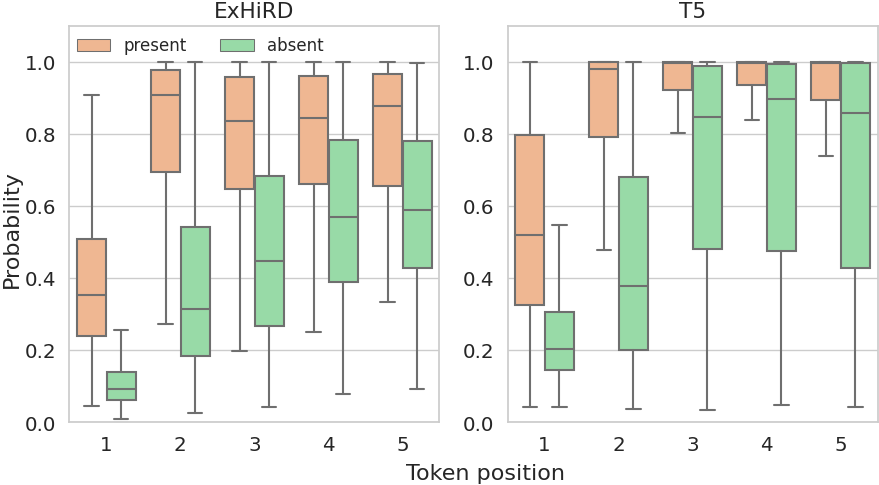}
  \caption{ExHiRD and T5's conditional probabilities for the first five tokens generated in a keyphrase (present and absent) in accordance to their relative positions within the keyphrase on the KP20K test set.} 
 \label{boxplots}
 \vspace{-4mm}
\end{figure}

In Figure \ref{boxplots}, we show that the {\em conditional probabilities of tokens in a keyphrase} tend to be low at the boundaries (at the beginning of a keyphrase), but start to increase monotonically as the decoder moves towards the end of the keyphrase. Intuitively, it makes sense that a model will have less confidence predicting the start of a keyphrase because it requires settling on a specific keyphrase to generate out of many potential candidates. However, the first keyphrase token, once already generated, will condition and restrict the space of plausible candidates for the second token thereby increasing its confidence. For the same reason, probabilities near the end of a keyphrase tend to be much higher. 
\begin{figure}[t]

 \center

  \includegraphics[width=\linewidth]{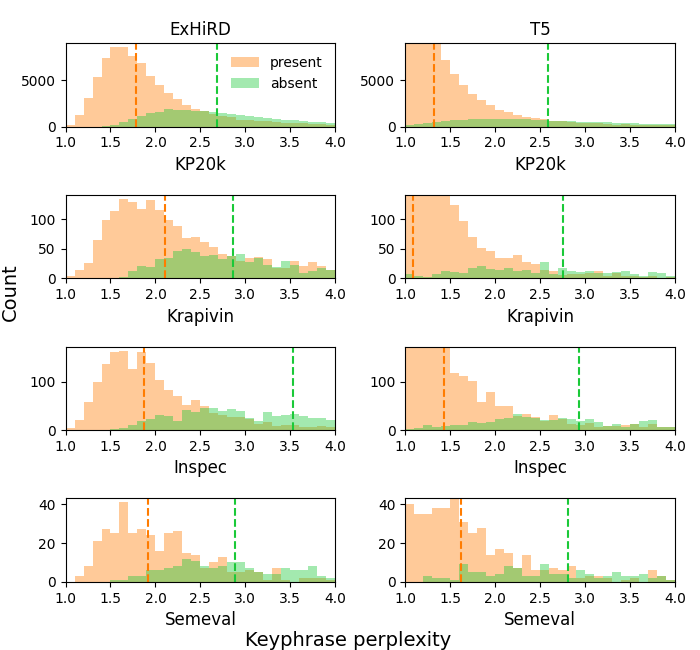}

  \caption{Histograms depicting number of keyphrases with certain keyphrase perplexity values for present and absent keyphrase generation. Dashed lines indicate the median of each distribution.}
  
  \label{confidence}
\vspace{-3mm}
\end{figure}

\begin{table}[b]

\centering
\small
\begin{tabular}{c|cc}
\toprule
Dataset & ExHiRD & T5 \\ \midrule
Inspec & 9.99 & 26.75 \\
Krapivin & 9.11 & 58.86 \\
SemEval & 10.18 & 26.64 \\
KP20k & 13.32 & 36.97 \\ \bottomrule
\end{tabular}
\caption{Expected calibration error (ECE) for ExHiRD and T5 on various datasets. T5's calibration is worse than ExHiRD (lower the better).}
\label{table:calibration}
\end{table}
\vspace{1mm}

\begin{figure}[t]

\centering

 \center
  \includegraphics[width=\linewidth]{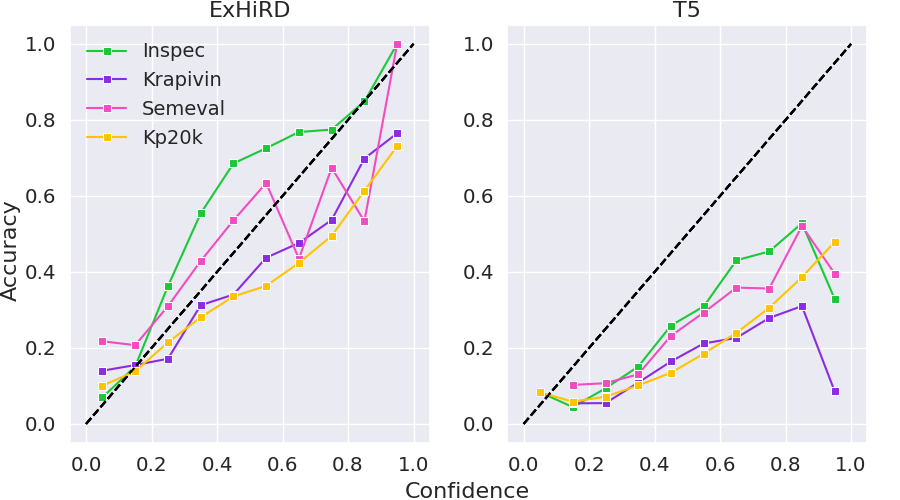}
  \caption{Reliability diagrams for model calibration of ExHiRD and T5. Dotted black line depicts perfectly calibrated model. We can see that ExHiRD is better calibrated than T5.}
  \label{Positional range}
  \label{fig:calibration}
  \vspace{-4mm}
\end{figure}

\noindent \textbf{Model Calibration}
In Figure \ref{confidence}, we saw that T5 predicts keyphrases with higher model confidence than ExHiRD (or CatSeq-Transformer). But does the higher confidence actually translate into better predictions? Figure \ref{fig:calibration} shows the reliability diagrams for ExHiRD and T5 for both present and absent keyphrases. We can see that calibration of ExHiRD is better than T5. T5's high confidence keyphrase predictions does not translate into optimal accuracy values. In Table \ref{table:calibration}, T5's ECE is much higher than ExHiRD for all four datasets. We can say that T5 is an \textit{overconfident} model. 
In Appendix \ref{section:vanilla_transformer_comparison} (Table \ref{table:one2seq_calibration}), we find that CatSeq-Transformer is also generally better calibrated than T5, but ExHiRD seems overall the best calibrated among all. 

\begin{table*}[h]
\small
\centering
\begin{tabular}{l|cccc|cccc}
\toprule
\multicolumn{1}{l|}{\multirow{2}{*}{Metric}} & \multicolumn{4}{c|}{ExHiRD} & \multicolumn{4}{c}{T5} \\
\multicolumn{1}{c|}{} & Inspec & Krapivin & Semeval & \multicolumn{1}{c|}{KP20k} & Inspec & Krapivin & Semeval & KP20k \\ \hline
\multicolumn{9}{c}{Present keyphrases} \\ \midrule
F$_1$@5 & 0.253 & \textbf{0.286} & \textbf{0.284} & 0.311 & \textbf{0.287} & 0.271 & 0.275 & \textbf{0.335}\\
F$_1$@M & 0.291 & \textbf{0.347} & \textbf{0.335} & 0.374 & \textbf{0.340}  & 0.328 &  0.306 &  \textbf{0.387}\\
\multicolumn{1}{l|}{F$_{KMR}$@M} & 0.366 & \textbf{0.366} & \textbf{0.393} & 0.408 & \textbf{0.392} & 0.347 & 0.349 & \textbf{0.415} \\
\multicolumn{1}{l|}{F$_{BS}$ (DeBERTa)@M} & 0.388 & \textbf{0.370} & \textbf{0.396} & 0.428 & \textbf{0.405} & 0.344 & 0.359 & \textbf{0.433} \\
\multicolumn{1}{l|}{F$_{BS}$ (RoBERTa)@M} & 0.442 & \textbf{0.434} & \textbf{0.467} & 0.459 & \textbf{0.459} & 0.414 & 0.464 & \textbf{0.466} \\
\multicolumn{1}{l|}{F$_{BS}$ (SciBERT)@M} & \textbf{0.588} & \textbf{0.572} & \textbf{0.528} & 0.588 & 0.587 & 0.550 & 0.490 & \textbf{0.589} \\ \hline
\multicolumn{9}{c}{Absent keyphrases} \\ \midrule
F$_1$@5 & 0.011 & 0.022 & \textbf{0.017} & 0.016 & \textbf{0.014} & \textbf{0.028} & 0.016 & \textbf{0.018}\\
F$_1$@M & 0.022 & 0.043  & \textbf{0.025}  & 0.032 & \textbf{0.025} & \textbf{0.053} & 0.023 & \textbf{0.036} \\
\multicolumn{1}{l|}{F$_{KMR}$@M} & 0.042 & \textbf{0.076} & \textbf{0.042} & \textbf{0.054} & \textbf{0.049} & 0.071 & 0.040 & \textbf{0.054} \\
\multicolumn{1}{l|}{F$_{BS}$ (DeBERTa)@M} & 0.049 & \textbf{0.088} & \textbf{0.044} & 0.065 & \textbf{0.067} & 0.081 & 0.042 & \textbf{0.067} \\
\multicolumn{1}{l|}{F$_{BS}$ (RoBERTa)@M} & 0.072 & \textbf{0.135} & \textbf{0.087} & 0.083 & \textbf{0.089} & 0.122 & 0.086 & \textbf{0.087} \\
\multicolumn{1}{l|}{F$_{BS}$ (SciBERT)@M} & 0.160 & \textbf{0.253} & \textbf{0.128} & 0.173 & \textbf{0.187} & 0.212 & 0.117 & \textbf{0.182} \\ \bottomrule
\end{tabular}
\caption{Exact match F$_1$ and SoftKeyScore-based performance of present and absent keyphrase generation. We use F$_{BS}$ to represent F$_{BERTScore}$. @5 metrics only keeps the top $5$ keyphrase predictions (following \citet{chen2020exclusive}, dummy keyphrases were added if there were $< 5$ predictions). @M metrics use the full model prediction for evaluation. For each metric we bold the best score among the two models.}
\vspace{-2mm}
\label{table:bertscore}
\end{table*}

\begin{figure}[!ht]
 \center
  \includegraphics[width=\linewidth]{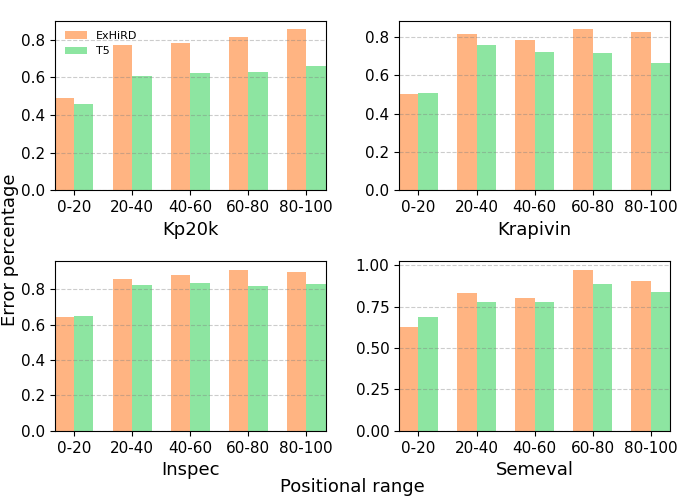}
  \caption{Error percentage of present keyphrase generation with respect to their position in the original text.}
  \label{positionalrange}
\vspace{-4mm}
\end{figure}

\vspace{1mm}
\noindent \textbf{Robustness to Positional Variance:}
In Figure \ref{positionalrange}, we find that both ExHiRD and T5 progressively get worse at identifying keyphrases in the later sections (bins) of the input document. However, ExHiRD's performance starts to fall more rapidly with increasing positions of the present keyphrases. In Appendix \ref{section:vanilla_transformer_comparison} (Figure \ref{one2seq_positionalrange}), we find that performances of T5 and CatSeq-Transformer degrades at a near similar rate. Thus, it appears that the Transformer architecture in general could be more robust in identifying keyphrases in different positions compared to RNN-based models. This could be the effect of non-local attention-based interactions in Transformers. 

 \vspace{1mm}
\noindent \textbf{Evaluation with Partial-match-based F1:} In Table \ref{table:bertscore}, we evaluate our models with different variations of SoftKeyScore (with KMR and variations of BERTScore as the \textit{score} function) and compare them with exact-match F$_1$ metrics. As expected, the magnitude of SoftKeyScore values are typically much higher that hard exact-match F$_1$ metrics. This is because predictions similar but slightly different from any of the ground truth keyphrases would be assigned a match score of $0$ in exact-match $F_1$, but can be given some value $\ge 0$ in SoftKeyScore. However, the two types of metrics roughly correlate, i.e., if T5 is better than ExHiRD in a dataset in F$_1$@M, we generally find it is also better in the SoftKeyScore metrics. Nevertheless, the difference of performance can vary widely based on the metric - for example, there is only a $.001$ absolute difference between T5 and ExHiRD on KP20K when using F$_{BS}$(SciBERT)@M but a difference of $.013$ when using F$_1$@M. More interestingly, sometimes, the worse model in the hard $F_1$-based metrics can become the better model in SoftKeyScore-based metrics. For example, under exact-match F$_1$, ExHiRD performs worse than T5 for absent keyphrases in Krapivin but under SoftKeyScores, ExHiRD performs better. Thus, SoftKeyScores do not always tell the same story as F$_1$.

\begin{table}[h]
\small
\centering
\begin{tabular}{lc}
\toprule
\multicolumn{1}{l|}{Metric} & Metric $\leftrightarrow$ Human \\ \midrule
\multicolumn{1}{l|}{F$_{1}$@M} & 0.3664 \\
\multicolumn{1}{l|}{F$_{KMR}$@M} & 0.4033 \\
\multicolumn{1}{l|}{F$_{BS}$ (DeBERTa)@M} & 0.3910 \\
\multicolumn{1}{l|}{F$_{BS}$ (RoBERTa)@M} & 0.3854 \\
\multicolumn{1}{l|}{F$_{BS}$ (SciBERT)@M} & 0.3543 \\ \bottomrule
\end{tabular}
\caption{Pearson correlation for various metrics against human scores of sets of predicted and gold keyphrases.}
\label{pearson_document}
\end{table}

\vspace{1mm}
\noindent \textbf{Human evaluation:}
To assess the quality of predicted keyphrases we use help from a CS majoring student. The student was asked to provide an appropriate score to signify the closeness between the predicted set of keyphrases and the gold set of keyphrases in $[0,1]$. The student was made familiar to the keyphrase generation task beforehand for better quality assessment and was instructed to look up unknown concepts or discuss them with the researchers of this study. Several iterations with the student were done before the actual annotation started.
The student compared and scored T5 prediction sets and the corresponding gold sets of $500$ sample documents from the KP20k test dataset. To ensure the quality of annotations, one of the authors of this paper evaluated the student's annotations for 25\% of the 500 documents. 

In Table \ref{pearson_document}, we show the Pearson correlation between various metrics when compared against the human scores. We see that all SoftKeyScore metrics except the one using SciBERT-based BERTScore are better correlated with human judgment than the F1 metric. Interestingly, F$_{BS}$(SciBERT)@M has the worst correlation. We find that SciBERT is generally more generous (overly-optimistic) with the magnitude of its similarity score than the other metrics whereas the human judgment is on a more conservative (realistic) side. Thus, SciBERT did not align well with the human evaluation. F$_{KMR}$@M, which is generally more conservative in its scoring, has the best correlation with the human evaluation. However, F$_1$@M is too conservative because even a minor difference in two keyphrases (predicted and gold) would imply a match score of 0. 

\section{Related Work}
\noindent \textbf{Keyphrase Generation:}
The current focus of research on keyphrase generation has been increasingly shifting towards seq2seq models particularly because of their capability to generate absent keyphrases \cite{meng2017deep}. Multiple works built upon seq2seq architectures to address keyphrase generation \cite{meng2017deep, chen-etal-2018-keyphrase, chan2019neural, chan-etal-2019-neural, swaminathan-etal-2020-preliminary, chen2020exclusive, ye-etal-2021-one2set, ye-etal-2021-heterogeneous, Huang_Xu_Jiao_Zu_Zhang_2021} (inter alia). Some recent works also explored the inclusion of pre-trained models for both absent and present keyphrase generation \cite{liu2020keyphrase, wu-etal-2021-unikeyphrase, mayank-learning, wu-constrained}. Our focus, however, is more in the analysis and evaluation rather than development of a new architecture. In terms of analysis, \citet{meng2020empirical} showed the effects of different hyperparameters including the ordering format for concatenating target keyphrases on the task. \citet{boudin2020keyphrase, boudin-gallina-2021-redefining} analyzed the contribution of present keyphrases and different types of absent keyphrase for document retrieval. 

\noindent \textbf{Model Calibration:}
Calibration and uncertainty of neural models \cite{guo2017calibration} have started to gain attention on several natural language processing tasks, including neural machine translation \cite{muller2019does, kumar2019calibration, wang-etal-2020-inference}, natural language understanding \cite{desai-durrett-2020-calibration}, coreference resolution \cite{nguyen-oconnor-2015-posterior}, and summarization \citet{xu2020understanding}. 
We analyze models for the keyphrase generation task on similar lines, as there is no previous literature available. 

\noindent \textbf{Evaluation metrics:}
In most NLG tasks (eg. summarization or machine translation) almost any evaluation is based on non-exact match scores (usually based on n-gram overlaps \cite{papineni-etal-2002-bleu, lin-2004-rouge}, or embedding-based distances \cite{zhang2019bertscore, sellam-etal-2020-bleurt}. However, 
they are not suitable to compare a set of phrases with another set of phrases as needed to be done in keyphrase generation. Instead, they are usually designed to check whether, on average, predicted sequences match very well with one of the reference sequences. In contrast, in keyphrase generation, we also need to account for coverage of all the reference keyphrases (we need to check whether most of the reference keyphrases are also highly matched by some prediction) among other factors. For instance, an ideal keyphrase evaluation metric should also account for undergeneration or overgeneration of keyphrases and take into account the structure of keyphrases, for example, by ignoring the order of keyphrases.

Among prior works in keyphrase generation and extraction, \citet{cano-bojar-2019-keyphrase} explored the use of summarization metrics such as ROUGE \cite{lin-2004-rouge} to compare concatenated sequences of keyphrases. While this can allow some partial match, this approach completely ignores the set-like structure of keyphrases. \citet{chan-etal-2019-neural} used Wikipedia information to control some level of name-variation over keyphrases of the same meaning but they still rely on strict binary scoring. \citet{luo2021keyphrase} devised a metric (to consider factors such as under-/over-generation) based on some partial-scoring functions but primarily to provide high quality reward signals in an RL setting. 

In contrast to them, our SoftKeyScore is a natural generalization of $F_1$ (which is already a dominant metric in keyphrase generation) to accounts for both partial-match and exact-match scoring functions. Moreover, as a benefit of its $F_1$ structure, SoftKeyScore can naturally and elegantly account for set to set comparisons, and penalize both undergeneration (with low recall) and overgeneration (with low precision) of keyphrases.

\section{Conclusion and Discussion}

In this work, we analyze, and compare three seq2seq models for keyphrase generation---a (non-pretrained) RNN-based model (ExHiRD), a (non-pretrained) Transformer Seq2Seq (CatSeq-Transformer), and a massively pre-trained Transformer-based model (T5) on model calibration, our newly proposed keyphrase perplexity, and robustness to positional variance of present keyphrases. Moreover, we propose a novel evaluation framework (SoftKeyScore) as a relaxed generalization of the standard F$_1$ metrics for evaluating keyphrase generation performance using soft-matching functions.

\section{Limitations}

Our analysis showcases key parameters of comparison between models in terms of KPP, calibration and positional variance for the keyphrase generation task. This provides insights into intrinsic model behavior while generating keyphrases. As we discussed before, one limitation of our $KPP$ measure as used in the study is that in a CatSeq framework, it is difficult to negate the effect of previously generated keyphrases. However, the keyphrase delimiters may naturally, to an extent, reduce the effect of previous keyphrases. Thus, it still can be decent heuristics. Not that Non-exact (quasi-)perplexity measures (in different formulations) have been also proposed in other contexts \cite{wang-etal-2019-make} before. There is also scope for further studies, for example, in investigation the effect of training with different keyphrase orders on positional robustness, or designing more accurate models to predict keyphrases that appear later in a document.  

\section{Ethics Statement}
We analyze various aspects of the keyphrase generation task. Keyphrase generation is a popular and established NLP task that is useful in information extraction. We do not forsee any ethical concern regarding our contribution to this domain.

\section{Acknowledgements}
This research is supported in part by NSF CAREER award \#1802358, NSF CRI award \#1823292, NSF IIS award \#2107518, and UIC Discovery Partners Institute (DPI) award. Any opinions, findings, and conclusions expressed here are those of the authors and do not necessarily reflect the views of NSF or DPI. We thank AWS for computational resources used for this study. 

\bibliography{anthology,custom}
\bibliographystyle{acl_natbib}

\clearpage
\appendix


\section{Implementation Details}
\label{eval}
ExHiRD is trained from the publicly available code \footnote{\url{https://github.com/Chen-Wang-CUHK/ExHiRD-DKG}} using the original settings mentioned in the paper \cite{chen2020exclusive}. CatSeq-Transformer is trained from the code\footnote{\url{https://github.com/jiacheng-ye/kg_one2set}} made publicly available by \citet{ye-etal-2021-one2set}.  T5 was trained with SM3 optimizer \cite{NEURIPS2019_8f1fa019} for its memory efficiency. We use a learning rate ($lr$) of $0.1$ and a warm up for $2000$ steps with the following formulation:
\begin{math}
    lr = lr \cdot minimum\left(1, \left(\frac{steps}{warmup\_steps}\right)^2\right)
\end{math}
The learning rate was tuned among the following choices: $[1.0, 0.1, 0.01, 0.001]$ (using grid search).
We use an effective batch size of $64$ based on gradient accumulation. We train T5 for $10$ epochs with a maximum gradient norm of $5$. Both models were trained using teacher forcing. We use train, validation and test splits from \citet{meng2017deep}. Following \cite{meng2019ordermatters, chen2020exclusive}, the keyphrases in the target sequence are ordered according to their position of first occurrence within the source text. The first occurring keyphrase in the source text appears first in the target sequence. The absent keyphrases were appended in the end according to their original order. Both T5 and ExHiRD experienced target sequences in that order during training. 
Predictions for both the models were generated through greedy decoding. We use a maximum length of $50$ tokens for T5 during decoding. We use a single NVIDIA V100 GPU for training and testing all our models except CatSeq-Transformer.  CatSeq-Transformer was trained in NVIDIA RTX A5000. 

\section{F$_1$ Evaluation Details}
\label{evaluation}
We used similar post-processing for evaluation as \citet{chen2020exclusive}. Concretely, we stemmed both target keyphrases and predicted keyphrases using Porter stemmer. We removed all duplicates from predictions after stemming. We determined whether a keyphrase is present or not by checking the stemmed version of the source document. For F$_1@M$ metrics we select all the keyphrase predictions generated by the model. For F$_1@5$, following \citet{chen2020exclusive}, if there were less than $5$ predictions, we append incorrect keyphrases to the predictions to make it exactly $5$. For SoftKeyScores, similar to F$_1@M$ we select all predicted keyphrases. However $@5$ variants of SoftKeyScores can be used as well. 

\section{SoftKeyScore Implementation}
\label{section:bertscore_appendix}

When we use KMR, we first stem the phrases being compared with Porter Stemmer.
We use the BERTScore implementation provided by the authors \footnote{\url{https://github.com/Tiiiger/bert_score}}. We use variations of pre-trained transformer model weights to compute BERTScore such as {\fontfamily{qcr}\selectfont microsoft/deberta-large-mnli} for DeBERTa, 
{\fontfamily{qcr}\selectfont roberta-large} for RoBERTa and {\fontfamily{qcr}\selectfont scibert-scivocab-uncased} for SciBERT. All the weights are streamlined and made available by \citet{wolf2020transformers}.  We also use baseline rescaling of BERTScore as done by \citet{zhang2019bertscore}. For both BERTScore and KMR based scoring functions, also use a threshold $t$ of $0.4$ such that the output of the score function becomes $0$ if it is $< t$. This makes prevent inflation of the overall score from low scoring matches.


\section{Comparison with vanilla transformer}
\label{section:vanilla_transformer_comparison}

In this section, we make comparisons with a vanilla transformer model, namely, the CatSeq-Transformer \cite{yuan-etal-2020-one}. The intuition behind comparing ExHiRD and T5 with CatSeq-Transformer is to see how does the behavior of a vanilla transformer model compare in comparison to a pre-trained transformer model and an RNN based model for the keyphrase generation task.

\begin{figure*}[!ht]
    \centering
    \subfigure[]{\includegraphics[width=0.48\linewidth]{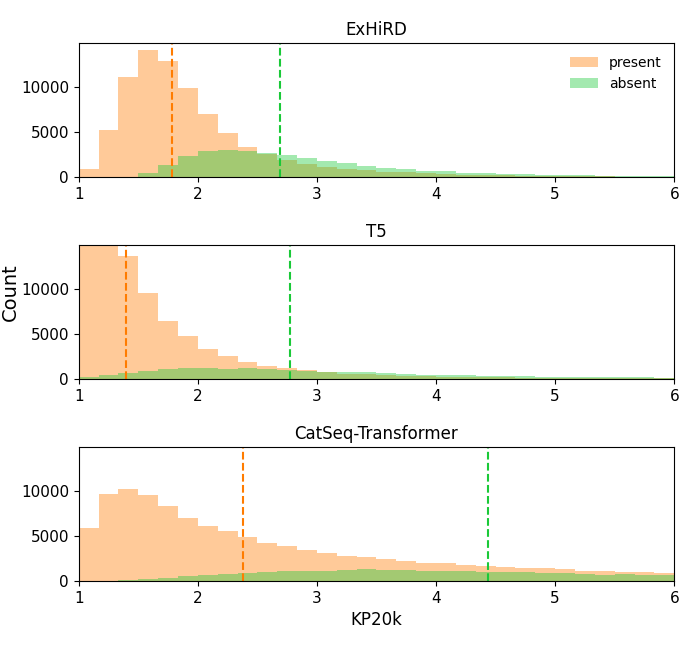}} 
    \subfigure[]{\includegraphics[width=0.48\linewidth]{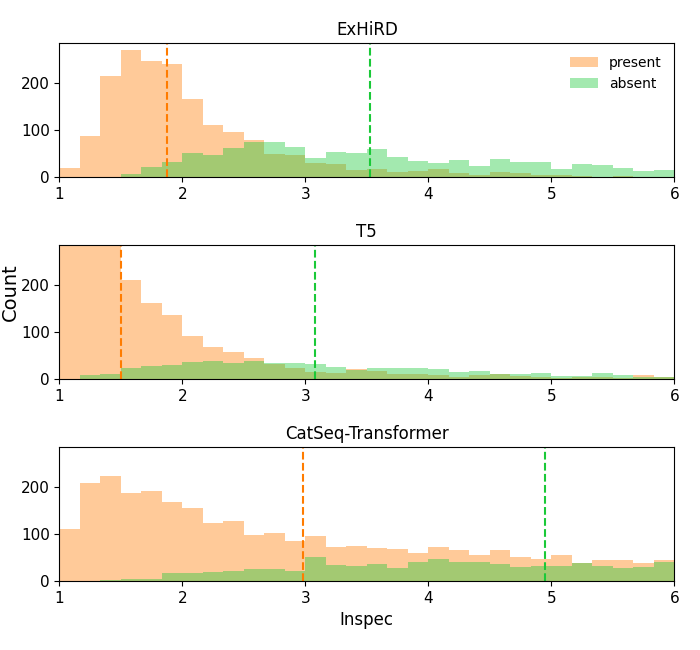}} 
    \subfigure[Keyphrase perplexity (KPP)]{\includegraphics[width=0.48\linewidth]{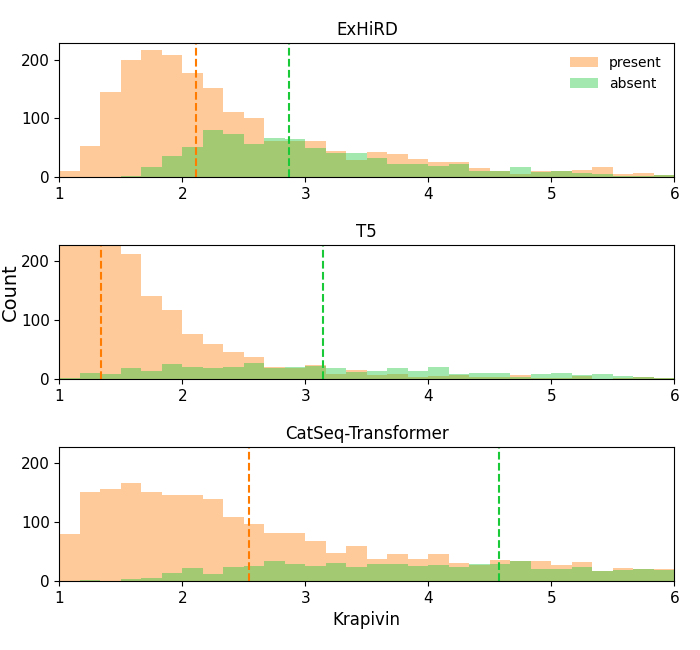}}
    \subfigure[Keyphrase perplexity (KPP)]{\includegraphics[width=0.48\linewidth]{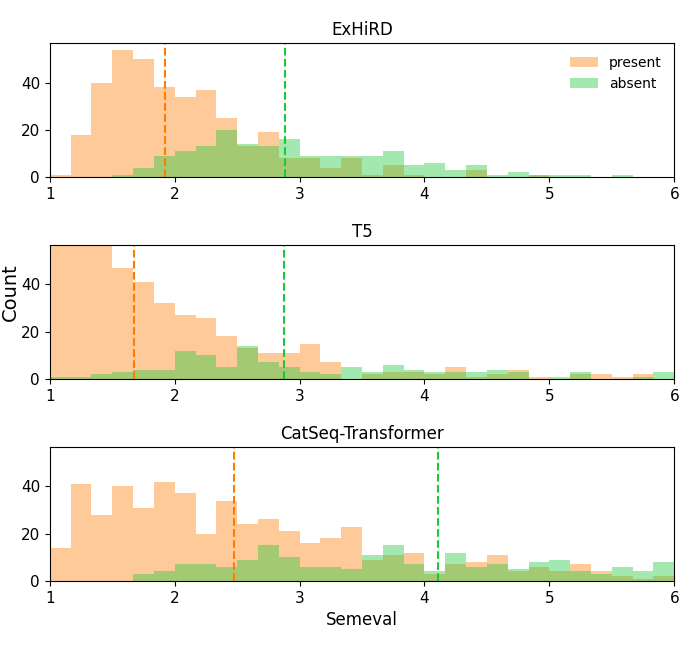}}
    
    \caption{Histograms depicting number of keyphrases in keyphrase perplexity bins of size 0.1 for present and absent keyphrase generation for (a) KP20k (b) Inspec (c) Krapivin (d) Semeval datasets. Dashed lines indicate the median of each distribution.}
    \label{fig:all_three_perplexities}
\end{figure*}

\begin{figure}[t]
 \center
  \includegraphics[width=\linewidth]{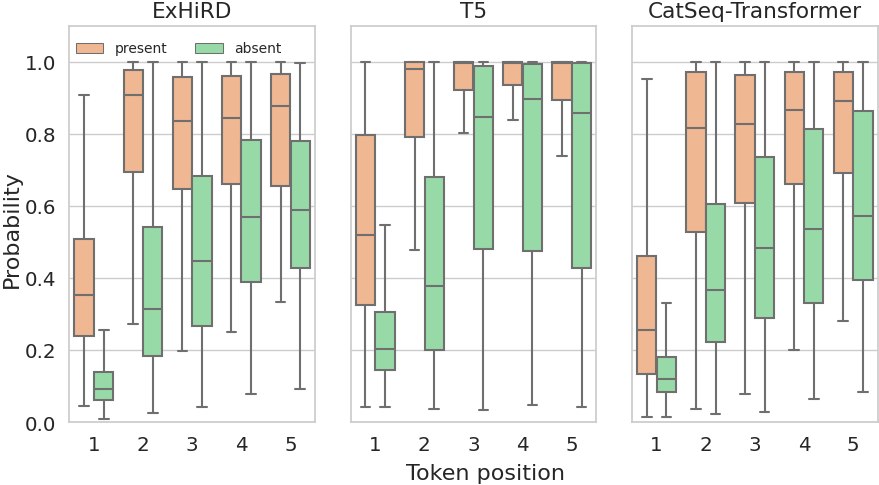}
  \caption{ExHiRD, T5 and CatSeq-Transformer's conditional probabilities for the first five tokens generated in a keyphrase (present and absent) in accordance to their relative positions within the keyphrase on the KP20K test set.} 
 \label{one2seq_boxplots}

\end{figure}

\vspace{1mm}
\noindent \textbf{Keyphrase Perplexity} In Figure \ref{fig:all_three_perplexities}, we compare the KPP of all three models using histograms that show KPP and the total number of keyphrases with certain KPP values. We notice that T5 has higher model confidence while generating present keyphrases than both ExHiRD and CatSeq-Transformer which could be the effect of model pre-training. Across all three models, we observe consistency in the fact that model confidence of absent keyphrases is lower than present keyphrases. In Figure \ref{one2seq_boxplots}, the conditional probabilities generated by the CatSeq-Transformer model for the first five tokens of the sequence are similar to ExHiRD, whereas T5 has higher values amongst all three models. 

\begin{figure}[]

\centering

 \center
  \includegraphics[width=\linewidth]{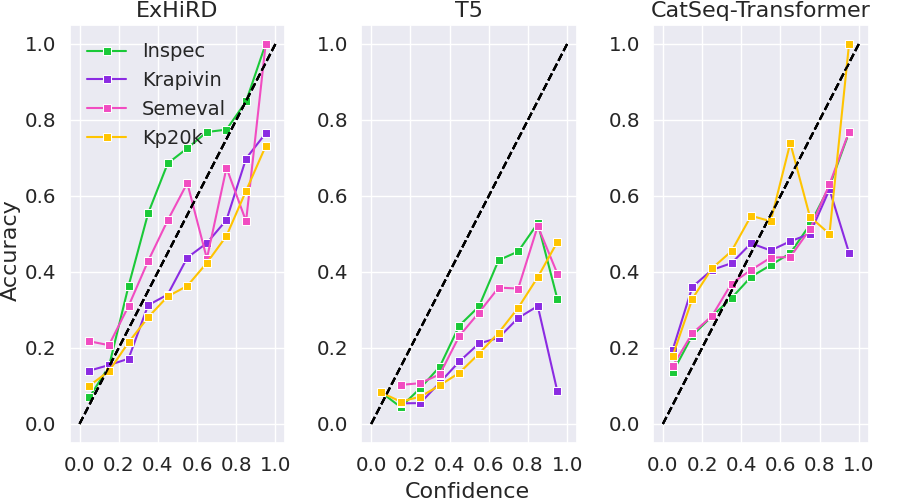}
  \caption{Reliability diagrams for model calibration of ExHiRD, T5 and CatSeq-Transformer. Dotted black line depicts perfectly calibrated model. We can see that ExHiRD is better calibrated than T5, but similar to CatSeq-Transformer}
  \label{Positional range}
  \label{one2seq_calibration}
\end{figure}

\begin{table}[!ht]

\centering
\small
\begin{tabular}{c|ccc}
\toprule
Dataset & ExHiRD & CatSeq-Transformer & T5 \\ \midrule
Inspec & 9.99 & 28.34 & 26.75 \\
Krapivin & 9.11 & 17.51 & 58.86 \\
SemEval & 10.18 & 20.73 & 26.64 \\
KP20k & 13.32 & 15.20 & 36.97 \\ \bottomrule
\end{tabular}
\caption{Expected calibration error (ECE) for ExHiRD and T5 on various datasets. T5's calibration is worse than ExHiRD and CatSeq-Transformer(lower the better).}
\label{table:one2seq_calibration}
\end{table}

\noindent \textbf{Calibration}
Figure \ref{one2seq_calibration} showcases CatSeq-Transformer's calibration to be comparatively similar to ExHiRD, whereas T5's reliability diagram highlights towards poor calibration. Table \ref{table:one2seq_calibration} also corresponds to high ECE for T5 over the two model, indicating poor calibration (lower the better).

\begin{figure}[!ht]
 \center
  \includegraphics[width=\linewidth]{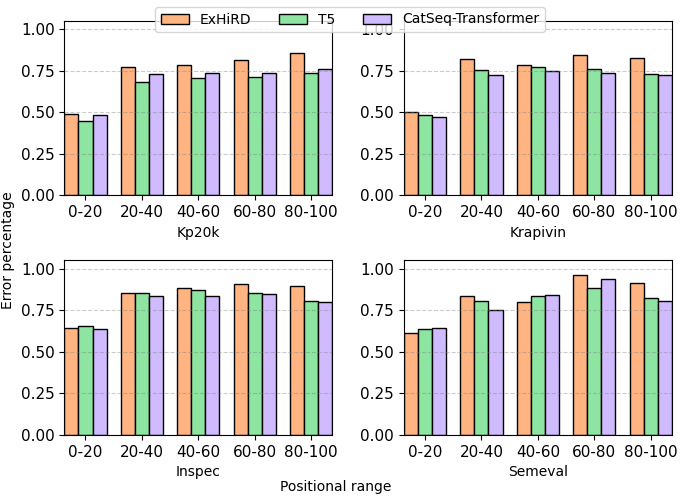}
  \caption{Error percentage of present keyphrase generation with respect to their position in the original text for ExHiRD, T5 and CatSeq-Transformer}
  \label{one2seq_positionalrange}
\end{figure}

\vspace{1mm}
\noindent \textbf{Positional variance}
As seen in Figure \ref{one2seq_positionalrange}, we observe ExHiRD has the higher error percentage towards the latter portions of the text in comparison to T5 and CatSeq-Transformer. It indicates that transformer based models are better at extracting keyphrases from the latter parts of the text.

\end{document}